\documentclass{article}

     \PassOptionsToPackage{numbers, compress}{natbib}

\usepackage[preprint]{neurips_2023}

\usepackage[utf8]{inputenc} 
\usepackage[T1]{fontenc}    
\usepackage{hyperref}       
\usepackage{url}            
\usepackage{booktabs}       
\usepackage{amsfonts}       
\usepackage{nicefrac}       
\usepackage{microtype}      
\usepackage{xcolor}         
\usepackage{microtype}
\usepackage{tablefootnote}
\usepackage{subcaption,graphicx}
\usepackage{booktabs} 
\usepackage{float}
\usepackage{hyperref}
\usepackage{array,multirow}

\usepackage[ruled, vlined, linesnumbered]{algorithm2e}
\usepackage{amsmath}
\usepackage{amssymb}
\usepackage{mathtools}
\usepackage{tabularx}
\usepackage[export]{adjustbox}
\usepackage{amsthm}
\usepackage{wrapfig}
\newcommand{\bs}{\boldsymbol}
\usepackage{amssymb}
\usepackage{pifont}
\usepackage{accents} 
\newcommand{\xmark}{\ding{55}}%
\newcommand{\asts}{{\boldsymbol{S}^{\ast}}}
\newcommand{\bss}{{\boldsymbol{S}}}
\newcommand{\mby}{{\mathbf{Y}}}
\newcommand{\mbty}{{\mathbf{\tilde Y}}}
\newcommand{\asty}{{\mathbf{Y}^*}}
\newcommand{\gty}{\mbf{Y}^{\text{gt}}}
\newcommand{\mc}{\mathcal}
\newcommand{\mbf}{\mathbf}

\theoremstyle{plain}

\theoremstyle{definition}

\theoremstyle{remark}

\usepackage[capitalize,noabbrev]{cleveref}

\title{ZegOT: Zero-shot Segmentation Through Optimal Transport of Text Prompts }

\author{%
 Kwanyoung Kim\thanks{equal contribution}, \quad  Yujin Oh$^{*}$,  \quad Jong Chul Ye \\
  School of AI, \\ 
  Korea Advanced Institute of Science and Technology (KAIST)\\
 Daejeon, Republic of Korea\\
 \texttt{\{cubeyoung, yujin.oh, jong.ye\}@kaist.ac.kr} \\
}

\begin{document}

\maketitle

\begin{abstract}

Recent success of large-scale Contrastive Language-Image Pre-training (CLIP) has led to great promise in zero-shot semantic segmentation by transferring image-text aligned knowledge to  pixel-level classification. 
However, existing methods usually require an additional image encoder or retraining/tuning the CLIP module. 
Here,  we propose a novel \textbf{Z}ero-shot s\textbf{eg}mentation with \textbf{O}ptimal \textbf{T}ransport (ZegOT) method that matches multiple text prompts with frozen image embeddings through optimal transport. In particular, we introduce a novel  Multiple Prompt Optimal Transport Solver (MPOT), which is designed to learn an optimal mapping between multiple text prompts and visual feature maps of the frozen image encoder hidden layers. This unique mapping method facilitates each of the multiple text prompts to effectively focus on distinct visual semantic attributes.
Through extensive experiments on benchmark datasets, we show that our method achieves the state-of-the-art (SOTA) performance over existing Zero-shot Semantic Segmentation (ZS3) approaches.  
\end{abstract}
\section{Introduction}
\label{submission}

Zero-shot Semantic Segmentation (ZS3) is one of label-efficient 
approaches for dense prediction task, which reduces efforts for expensive pixel-level annotations of unseen object categories \cite{bucher2019zero}.
Vision language models (VLM) such as CLIP \cite{radford2021clip} have brought great advance in  ZS3 task by transferring pre-trained image-text aligned knowledge to 
{pixel-text level} category matching problems \cite{ding2022zegformer, xu2021zsseg, zhou2022maskclip, rao2022denseclip, zhou2022zegclip}.  
The key idea here is to use the VLM knowledge gained through contrastive learning on large-scale image-text pairs through a special knowledge transfer process tailored to ZS3.


Suppose that our goal is to learn a functional map that transfers the pre-trained image-level domain $\mathcal{X}$ knowledge to the novel pixel-level domain $\mathcal{Y}$. 
To achieve this goal, various approaches have been explored by training additional models or leveraging a novel domain knowledge, 
as indicated in Table~\ref{table:intro}.
In general, existing ZS3 approaches based on pre-trained VLM can be categorized into two groups: Frozen image encoder with learnable proposal generator-based approaches (FE) \cite{ding2022zegformer, xu2021zsseg}, and trainable image encoder-based approaches (TE) covering re-training or fine-tuning \cite{rao2022denseclip, zhou2022zegclip}.





\begin{figure}[!t]
\centering
\includegraphics[width = 0.98\linewidth]{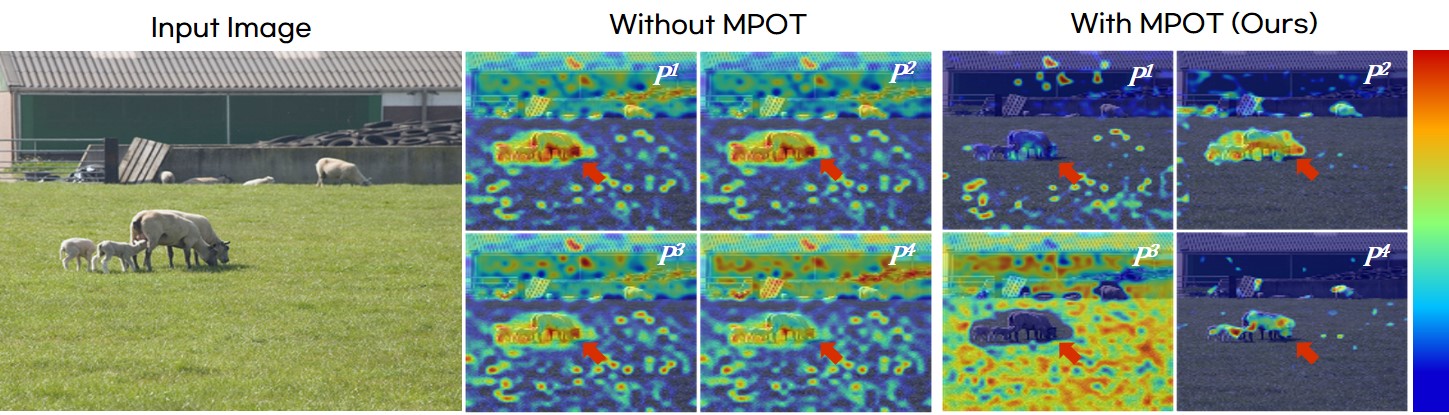}
\vspace{-0.1cm}
\caption{Visualization of text prompts-related image pixel alignment. Without proposed MPOT, {all the multiple prompts on each class ${P}^i$ are cohered} and their related score maps resemble each others. On the other hand, with our MPOT, each ${P}^i$ is diversely projected and each ${P}^i$-related score map focuses on different semantic attributes.
}
\label{fig_main}
\vskip -0.1in
\end{figure}


FE approaches exploit the $\mathcal{Y}$ domain knowledge through image information from the proposal generator. These approaches utilize the capability of image-level classification in the frozen image encoder.
A major drawback of FE approaches is that they require an additional trainable image encoder to generate region proposals, which leads to high-cost memory complexity. Also, FE performance is highly dependent on the frozen VLM, which is not tailored for pixel-level dense prediction.
On the other hand, TE approaches explicitly address the VLM dependency issue by directly training another image encoder for dense prediction and exploiting $\mathcal{Y}$ domain knowledge through class-driven text information.
Compared to FE approaches, leveraging novel domain knowledge through textual information is simple and efficient in terms of complexity.  
Since TE approaches are learned for a specific domain, they achieve superior performance compared to FE approaches for the corresponding ZS3 task.
However, TE approaches also suffer from high computational cost due to the learnable image encoder. 

\begin{wraptable}[13]{r}{0.65\textwidth}
\vspace{-1em}
\caption{Systemic analysis of ZegOT compared to CLIP-based segmentation models. OCL denote Orthogonal Constrained Loss. } 
\vspace{0em}
\label{table:intro}
\begin{center}
\resizebox{1\linewidth}{!}{
\begin{tabular}{lcccc}
\toprule
\multirow{2}{*}{Model}  & Proposal &  CLIP module & Multiple & Multi-prompts\\ 
& generator & retrain/fine-tune  & text prompt& pixel matching\\
\cmidrule(r){1-1} \cmidrule(lr){2-5}
ZegFormer~\cite{ding2022zegformer} &  \checkmark  & \xmark & \xmark & - \\
zsseg~\cite{xu2021zsseg}     &  \checkmark &  \xmark & \xmark  & - \\
MaskCLIP+~\cite{zhou2022maskclip} & \xmark   & \xmark & \xmark  & -\\
DenseCLIP~\cite{rao2022denseclip} & \xmark  &  \checkmark & \xmark & -\\
ZegCLIP~\cite{zhou2022zegclip}   & \xmark  &  \checkmark  & \xmark & - \\
Freeseg~\cite{qin2023freeseg}   & \xmark  &  \checkmark  & \checkmark & Cross-attention \\
MVP-SEG+~\cite{guo2023mvp}   & \xmark  &  \xmark  & \checkmark & OCL loss\\
\cmidrule(r){1-1} \cmidrule(lr){2-5}
\textbf{ZegOT (Ours)} &  \xmark  &  \xmark &  \checkmark    & Optimal Transport\\
\bottomrule
\end{tabular}
}
\end{center}

\end{wraptable}

One naive solution  could be letting the image decoder and text prompts as only learnable parts, while keeping
the VLM image encoder and text encoder frozen.
However, this solution is problematic for the dense prediction task because the {learnable multiple} text prompts and image embeddings are not effectively aligned since the VLM only utilizes global alignment between image-sentence pair. 
Furthermore, 
under the condition of utilzing the frozen VLM,
simply introducing the multiple text prompts is sub-optimal and the learned text prompts can be converged to represent similar semantic features {as demonstrated in Figure~\ref{fig_main}}.

Interestingly, we observe that our proposed MPOT, incorporating the concept of optimal transport (OT) theory, effectively addresses the above problem by enabling each text prompt to selectively focus on particular semantic features. This approach leads to yielding optimally aligned pixel-text score maps, as demonstrated in Figure~\ref{fig_main}.
{The aligned score map through OT assists the model to effectively attend to target object features, leading our ZegOT robust to segmentation performance when dealing with unseen classes.} 
More specifically, we initially optimize the optimal transport plan between multiple text prompts and the frozen image embeddings on seen classes, and then utilize the optimized transport plan to predict pixel-text score maps for both seen and unseen classes. This allows us to achieve the state-of-the-art (SOTA) performance on ZS3 tasks. 


Our contributions can be summarized as:
\begin{itemize}
\item We propose a novel framework, called ZegOT, that requires only {learnable} multiple text-prompts and a decoder part to train while keeping the entire VLM frozen, which allows our model to fully leverage highly aligned vision and language information for zero-shot semantic segmentation tasks.
\item 
In ZegOT,
the proposed multi prompt optimal transport solver module
{effectively} matches the distribution between image embeddings from the frozen VLM and the learnable text-prompts. 
\item {Through extensive experiments on three benchmark datasets, we demonstrate that our ZegOT achieves SOTA performance for zero-shot semantic segmentation tasks to the previous SOTA approach.} 
\end{itemize}


\begin{figure*}[!t]
\vskip 0.1in
\begin{center}
\includegraphics[width=0.97\linewidth]{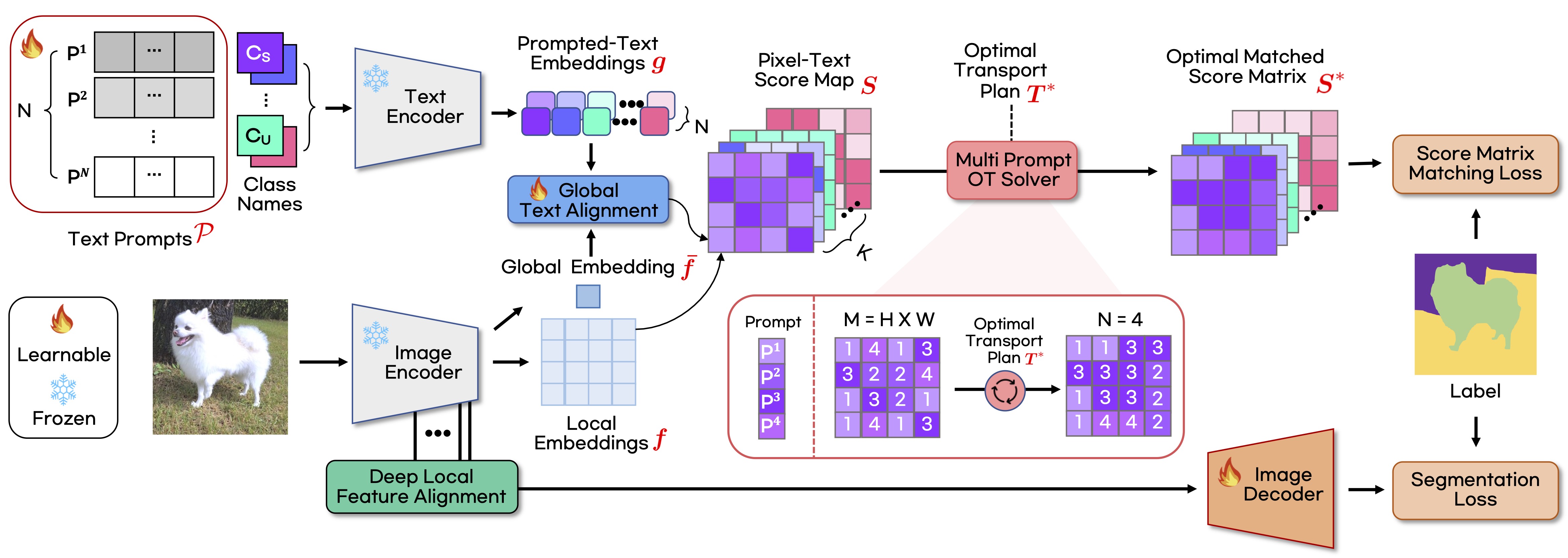}
\caption{Overall pipeline of our proposed ZegOT for zero-shot semantic segmentation. The only learnable parts are the multiple text prompts $\mathcal{P}$, the single linear layer for global text alignment, and image decoder while entire CLIP encoder modules are frozen.}
\label{fig_main0}
\end{center}
\vskip -0.1in
\end{figure*}


\section{Related Work}

\subsection{Prompt Learning}
Prompt learning has been initially introduced in the field of Natural Language Processing (NLP), which efficiently adapts large-scale model knowledge to various downstream tasks.
Rather than using traditional self-supervised learning paradigm,  \textit{i.e.,}  pre-train and fine-tune the large-scale model to transfer the knowledge to downstream tasks, 
the prompt learning formulates the downstream adjustment problem by training light-weight optimal textual prompts~\cite{petroni2019language,shin2020autoprompt,jiang2020can,liu2021gpt}. 
Compared to the fine-tuning method, text prompts-driven downstream adaptation is efficient, but still alleviates the domain shift problem that occurs between the pre-trained domain and the downstream domain. 
Recently, introducing a set of learnable text prompts into the frozen VLM achieved superior performance in various computer vision (CV) tasks~\cite{zhou2022learning, zhou2022conditional, gao2021clip,rao2022denseclip}. 
Visual Prompt Tuning (VPT) is also a novel solution that introduces trainable image prompts to each layer of a transformer encoder~\cite{jia2022visual}, 
which can be transferred to various downstream tasks~\cite{zang2022unified,sohn2022visual,liu2022prompt}. 
Our method utilizes multiple text prompts to adapt VLM to the segmentation task, while keeping the entire encoder part frozen. 

\subsection{Zero-shot Semantic Segmentation}
Semantic segmentation is a core computer vision task to densely analyze visual context. Recent success of CLIP \cite{radford2021clip} accelerates the advancement of language-driven semantic segmentation by utilizing pre-aligned vision-language knowledge between the image encoder and the class name-driven text encoder \cite{li2022language, xu2022groupvit, liang2022open}.
However, this dense prediction task requires a labor-intensive pixel-level annotation process to yield reliable performance. This leads a label-imbalance issue, \textit{i.e.,} not all the categories from the large vocabulary are annotated in the training dataset.
Zero-shot semantic segmentation (ZS3) solves this label-imbalance problem by generalizing labeled (seen) class knowledge to predict new (unseen) class information \cite{bucher2019zero}. 
MaskCLIP+ \cite{zhou2022zegclip} introduces a ZS3 method by simply extracting the text-driven visual features from the CLIP image encoder. 
ZegCLIP \cite{zhou2022zegclip} successfully bridges the performance gap between the seen and unseen classes by adapting a visual prompt tuning technique instead of fine-tuning the frozen CLIP image encoder. 
FreeSeg \cite{qin2023freeseg} introduces a generic segmentation framework by training task-adaptive text prompts. 
Recently, MVP-SEG+ \cite{guo2023mvp} employs Orthogonal Constraint Loss (OCL) to exploit CLIP features on different object parts.
However, our proposed ZegOT tries to extract CLIP knowledge by introducing multiple text prompts through a novel matching method, Multiple
Prompt Optimal Transport Solver (MPOT).

ZS3 can be performed by either inductive or transductive settings. Compared to traditional inductive zero-shot setting where class names and pixel-level annotations of unseen classes are both unavailable during training \cite{ding2022zegformer}, a newly introduced transductive setting boosts the ZS3 performance by utilizing unseen class names and self-generated pseudo labels guided by the model itself during training \cite{gu2020cagnet, xu2021zsseg, pastore2021strict, zhou2022maskclip, zhou2022zegclip}. {We adapt ZegOT to the newly introduced transductive ZS3 setting.}

\subsection{Optimal Transport}
Optimal transport (OT) is a general mathematical framework to evaluate correspondences between two distributions. 
Thanks to the luminous property of distribution matching, the optimal transport  has received great attention in various computer vision tasks, such as domain adaptation~\cite{flamary2016optimal}, semantic correspondence problem~\cite{liu2020semantic}, graph matching~\cite{xu2019scalable,xu2019gromov}, and cross-domain alignment~\cite{chen2020graph}, etc.  
Among various methods, Sinkhorn algorithm can efficiently solve the OT problem through entropy-regularization~\cite{cuturi2013sinkhorn}, and it can be directly applied to deep learning frameworks thanks to the extension of Envelop Theorem~\cite{peyre2019computational}. 
Prompt Learning with Optimal Transport (PLOT) \cite{chen2022prompt} is mostly related to ours, which optimizes the optimal transport distance to align visual features and text features by the Sinkhorn algorithm given trainable multiple text prompts. However, PLOT only considers a few-shot image classification problem, while we apply the optimal transport theory to a zero-shot dense prediction problem to focus on
specific attributes of semantic objects, which facilitates ZS3.  


\section{Preliminary}
\paragraph{Optimal Transport Problem}
Optimal transport aims to minimize the transport distance 
between two probability distributions. In this paper, we only consider discrete distribution which is closely related to our framework. We assume that discrete empirical distributions $\bs{\mu}$ and $\bs{\nu}$ are defined on probability space $\mathcal{F}, \mathcal{G} \in \Omega$, respectively, as follows:
\begin{align}
\bs{\mu} = \sum^{M}_{i=1} p_{i} \delta_{f_{i}}, \quad \bs{\nu} = \sum^{N}_{j=1} q_{j} \delta_{ g_{i}}, 
\end{align}
where $\delta_f$ and $\delta_g$ denote Dirac functions centered on $\bs{f}$ and $\bs{g}$, respectively, $M$ and $N$ denote the dimension of the empirical distribution. 
The weight vectors $\boldsymbol{p} = \{p_i\}^M_{i=1}$ and $\boldsymbol{q} = \{q_i\}^{N}_{j=1}$  belong to the $M$ and $N$-dimensional simplex, respectively, \textit{i.e.}, $\sum^{M}_{i=1} p_i = 1$, $\sum^{N}_{j=1} q_j = 1$. The discrete optimal transport problem  can be then formulated as: 
\begin{eqnarray}
\bs{T}^{\ast} = \underset{\bs{T}\in \mathbb{R}^{MXN}}{\arg{\min}} \sum^{M}_{i=1}\sum^{N}_{j=1}\bs{T}_{ij} \bs{C}_{ij} \nonumber \\ \textrm{s.t.} \quad \bs{T}\bs{1}^{N} = \bs{\mu}, \quad \bs{T}^{\top}\bs{1}^{M} = \bs{\nu} .
\label{DOT}
\end{eqnarray}
Here,
$\bs{T}^{\ast}$ is called the optimal transport plan, which is learned to minimize 
the total distance between the two probability vectors. $\bs{C}$ is the cost matrix which represents the distance between $\boldsymbol{f}_i$ and $\boldsymbol{g}_j$, \textit{e.g.,} the cosine distance $\bs{C}_{ij}$ = 1 - $\frac{\bs{f}_i\bs{g}^{\top}_j}{\|\bs{f}_i\|_2 \|\bs{g}_j\|_2}$, and $\bs{1}^{M}$ refers to the $M$-dimensional vector with ones. 

However, solving the problem~\eqref{DOT} costs $O(n^3\log n)$-complexity ($n$ proportional to $M$ and $N$), which is time-consuming. 
This issue can be efficiently solved by entropy-regularizing the objective, called the Sinkhorn-Knopp (or simply Sinkhorn) algorithm~\cite{cuturi2013sinkhorn}. In
Sinkhorn algorithm, the optimization problem is reformulated as:
\begin{eqnarray}
\bs{T}^{\ast} = \underset{\bs{T}\in \mathbb{R}^{MXN}}{\arg{\min}} \sum^{M}_{i=1}\sum^{N}_{j=1}\bs{T}_{ij}\bs{C}_{ij} - \lambda H(\bs{T}) \nonumber \\ \textrm{s.t.} \quad \bs{T}\bs{1}^{N} = \bs{\mu}, \quad \bs{T}^{\top}\bs{1}^{M} = \bs{\nu} .
\label{Sinkhorn}
\end{eqnarray}
where $H(\bs{T})$ = $\sum_{ij} \bs{T}_{ij} \log \bs{T}_{ij}$ and $\lambda > 0$ is the regularization parameter.
The problem \eqref{Sinkhorn} is a strictly convex optimization problem, and thus we have an optimization solution with fewer iterations as follows: 
\begin{eqnarray}
\bs{T}^{\ast} = \text{diag}(\bs{a}^{t})\exp(-\bold{C}/\lambda)\text{diag}(\bs{b}^{t})
\label{Sinkhorn2}
\end{eqnarray}
where $t$ is the iteration and $\bs{a}^t = \bs{\mu}/ \exp(-\bold{C}/\lambda)\bs{b}^{t-1}$ and $\bs{b}^{t} = \bs{\nu}/\exp(-\bold{C}/\lambda)\bs{a}^{t}$, with the initialization on $\bs{b}^{0}=\bs{1}$.

\section{Methods}
\label{methods}
As illustrated in Figure~\ref{fig_main0}, our proposed ZegOT is composed of frozen CLIP text and image encoder modules with trainable text prompts, a global text alignment (GA) layer, and a multi prompt optimal transport solver (MPOT) module. 
A primary goal of ZegOT is to segment objects belongs to both seen classes $\mc{C}_S$ and unseen classes $\mc{C}_U$, \textit{i.e.,} $\mc{C} = \mc{C}_{S} \cup \mc{C}_U$, where $\mc{C}_S \cap \mc{C}_U = \emptyset$, for the transductive ZS3 settings. 
For the transductive setting, the class names of $\mc{C}_U$ are further provided during training.
We introduce two novel approaches into ZegOT to learn dense vision-language alignment in a cost-effective way: 1) prompt-guided deep text-pixel alignment and 2) multi prompt optimal transport solver.

\subsection{Prompt-guided Deep Text-Pixel Alignment}


 To fully leverage the CLIP pre-trained knowledge, we make $N$ text prompts $\mathcal{P} = \{\bs{P}^{i}|^N_{i=1}\}$ as only trainable context tokens from the text encoder module. \textit{i.e.,} $\bs{P}^i =[P^i_1,P^i_2, \cdots, P^i_l],$ where $l$ denotes the length of the context tokens. 
 The randomly initialized multiple text prompts $\mathcal{P}$ are identically prepended to $K$ tokenized class names as $\mathcal{T} = \{\{\mathcal{P}, \bs{c}^k\}|^K_{k=1}\}$, where $\{\bs{c}^k|_{k=1}^K\} \subset \mc{C}$ is the word embedding of each class name. Note that the text prompts $\mathcal{P}$ are shared throughout all the class names.

Now,  an input image is encoded through the frozen CLIP image encoder layer to yield  the $i$-th intermediate local image embedding {$\bs{f}_i \in \mathbb{R}^{H_LW_L \times D}, i=1,\cdots, L$  from the $i$-th hidden image encoder layer, where $H_{L}$ and $W_{L}$ are the height and width of the local image embeddings from $L$-th layer and $D$ denotes the embedding dimension. 
Using these, we define variants of desirable pixel-text aligned score matrices as: 
\begin{align}
	\bs{S}_i = \bs{f}_i {\bs{g}^{\top}_{GA}},\quad  \asts_i= \bs{T}^{\ast}_i \odot\bss_i,
	\label{eq:score}
\end{align} 
where the superscript $^{\top}$ refers to the transpose operation, 
$\asts_{i} \in \mathbb{R}^{H_LW_L \times KN}$ is the refined score matrix by optimal transport map $\bs{T}^{\ast}_i$ which will be discussed in Sec~\ref{MPOT}. 
Furthermore, we define global image embedding $\bs{\bar{f}}_L \in \mathbb{R}^{1 \times D}$, and
$$ \bs{g}_{GA} = \mathcal{Q} (\text{cat}\left[\bs{\bar{f_L}}  \odot \bs{{g}}, \bs{{g}} \right] )  \in \mathbb{R}^{KN \times D}$$
 is the globally aligned text embedding where $\mathcal{Q}$ denotes a linear layer for matching the concatenated embedding dimension to the original dimension $D$,   \text{cat} is the concatenation operator,  and $\odot$ is the Hadamard product,
and  $\bs{g} = E_{\text{text}}(\mathcal{T}) \in {\mathbb{R}^{KN \times D}}$ denotes the 
text embeddings from the $N$-prompted class names $\mathcal{T}$, where $E_{\text{text}}$ is the frozen text encoder.
Finally, $\bs{f}_i,\bs{\bar{f}}_L,\bs{g}$ and $\bs{g}_{GA}$ are $\mathcal{L}_2$ normalized along the embedding dimension. 

%
%

The score matrices in $\eqref{eq:score}$ can be fed into the image decoder as an intermediate feature to predict segmentation maps or directly matched to the pixel-level annotation labels as:
\begin{align}
\mby = \mathcal{D}_{\theta}(\text{cat}[\mathcal{M}(\bss_i)|^{L}_{i=1}]), \quad \mbty = \mathcal{U}(\mathcal{M}(\asts)).
 \label{eq:predict3}
\end{align}
where $\mby \in \mathbb{R}^{HW \times K}$ and $\mbty \in \mathbb{R}^{HW \times K}$ are the output of decoder and the prediction of aligned score matrix, respectively, $\mathcal{D}_{\theta}$ is the trainable image decoder, and $H$ and $W$ are the height and width of image, respectively.  $\asts$ denotes the final combined score matrix by incorporating intermediate score matrices $\asts_i$, as will be explained in Section~\ref{MPOT} . 
$\mathcal{M}$: $\mathbb{R}^{H_LW_L \times KN} \rightarrow \mathbb{R}^{H_L W_L \times K}$ is the operation which first reshape $\mathbb{R}^{H_L W_L \times KN} \rightarrow \mathbb{R}^{H_L W_L \times K \times N}$ and perform summation the matrix along the $N$ dimension, and  
$\mathcal{U}:\mathbb{R}^{H_LW_L \times K} \rightarrow \mathbb{R}^{HW \times K}$ is the upsampling operator ($H_LW_L < HW$).

In order to synergistically exploit the collective knowledge derived from both general information encapsulated in $\mby$ and the optimally transported score matrix $\mbty$, we formulate the final segmentation output $\asty$ as follows:
\begin{align}
\asty = \lambda \cdot \mby  + (1-\lambda) \mbty
 \label{eq:predictfinal}
\end{align}
where $\lambda \in [0,1]$ denotes the hyper-parameter for controlling the balance between $\mby$ and $\mbty$. 
Accordingly, 
$\asty$ is the final prediction of ZegOT.

\subsection{Multi Prompt Optimal Transport Solver (MPOT)}
\label{MPOT}

\begin{algorithm}[h!]
	\caption{Multiple Prompt Optimal Transport Solver  with Sinkhorn algorithm}
	\label{algo-ot}
	\SetKwInOut{Input}{Input}
	\SetKwInOut{Output}{Output}
	\SetKwInput{kwInit}{Initialization}
	\SetKwInput{kwset}{Given}
	\kwset{The feature map size $M = H_LW_L$, the number of prompts $N$,  $\bs{\mu} = \bs{1}^{M}/M$ , $\bs{\nu} = \bs{1}^{N}/N$, the score matrix $\bss$ ;}
	\Input{The cost matrix $\bs{C} = \bs{1} - \bss$, hyper-paramter $\bs{\epsilon}$, the max iteration $t_{\text{max}}$;}
	\kwInit{$\bs{K} = \exp(-\bs{C}/\bs{\epsilon})$, $t \leftarrow 1, \bs{b}^0 = 1$;}
	\While{$t \leq t_{\text{max}}$ $\mathbf{and not converge}$ }{
	$\bs{a}^t = \bs{\mu}$ / $(\bs{Kb}^{t-1})$; \\ 
	$\bs{b}^t = \bs{\nu}$ / $(\bs{K^{\top}a}^t)$;
	}
	\Output{Optimal transport plan $\bs{T}^{\ast}$ = $\text{diag}(\bs{a})^t\bs{K}\text{diag}(\bs{b})^t$ ;}
\end{algorithm}
 \vspace{-1em}

To incorporate the OT theory into our framework, we define the total cost matrix $\bs{C}$ in ~\eqref{Sinkhorn} using the globally aligned score matrix $\bs{S}_i$ in~\eqref{eq:score}, \textit{i.e.,} for the $i$-th layer, we set 
{$\bs{C} :=\bs{C}_i  = \bs{1 - S}_i$,} where $\bs{C}_i \in \mathbb{R}^{H_LW_L \times K N}$ denotes the $i$-th cost matrix.

Given the cost matrix $\bs{C}_i$, the goal of MPOT is to obtain the corresponding optimal transport plan $\bs{T}^{\ast}_L  \in \{\bs{T}^{\ast}_i\}^L_{i=1}$ as 
as Eq.~\eqref{Sinkhorn2} (Here, $M = H_LW_L$, $N= N$ in \eqref{Sinkhorn})., which is a mapping matrix that maximizes the cosine similarity between the frozen image embedding and the learnable multiple prompts-derived text embeddings as depicted in Algorithm~\ref{algo-ot}.
Since our goal is to obtain a pixel-text score matrix that minimizes the total distance between the two image and text embedding spaces, we can reformulate the intermediate score matrix as : 
\begin{align}
\bs{S}^{\ast}_i= \bs{T}^{\ast}_i \odot\bs{S}_i
\label{eq:shallow}
\end{align}
where $\bs{S}^{\ast}_i$ is the refined score matrix with transport plan $\bs{T}^{\ast}_i$ in $i$-th layer. 

\begin{wrapfigure}[20]{r}{0.5\textwidth}
\vspace{0em}
  \begin{center}
    \includegraphics[width=0.47\textwidth]{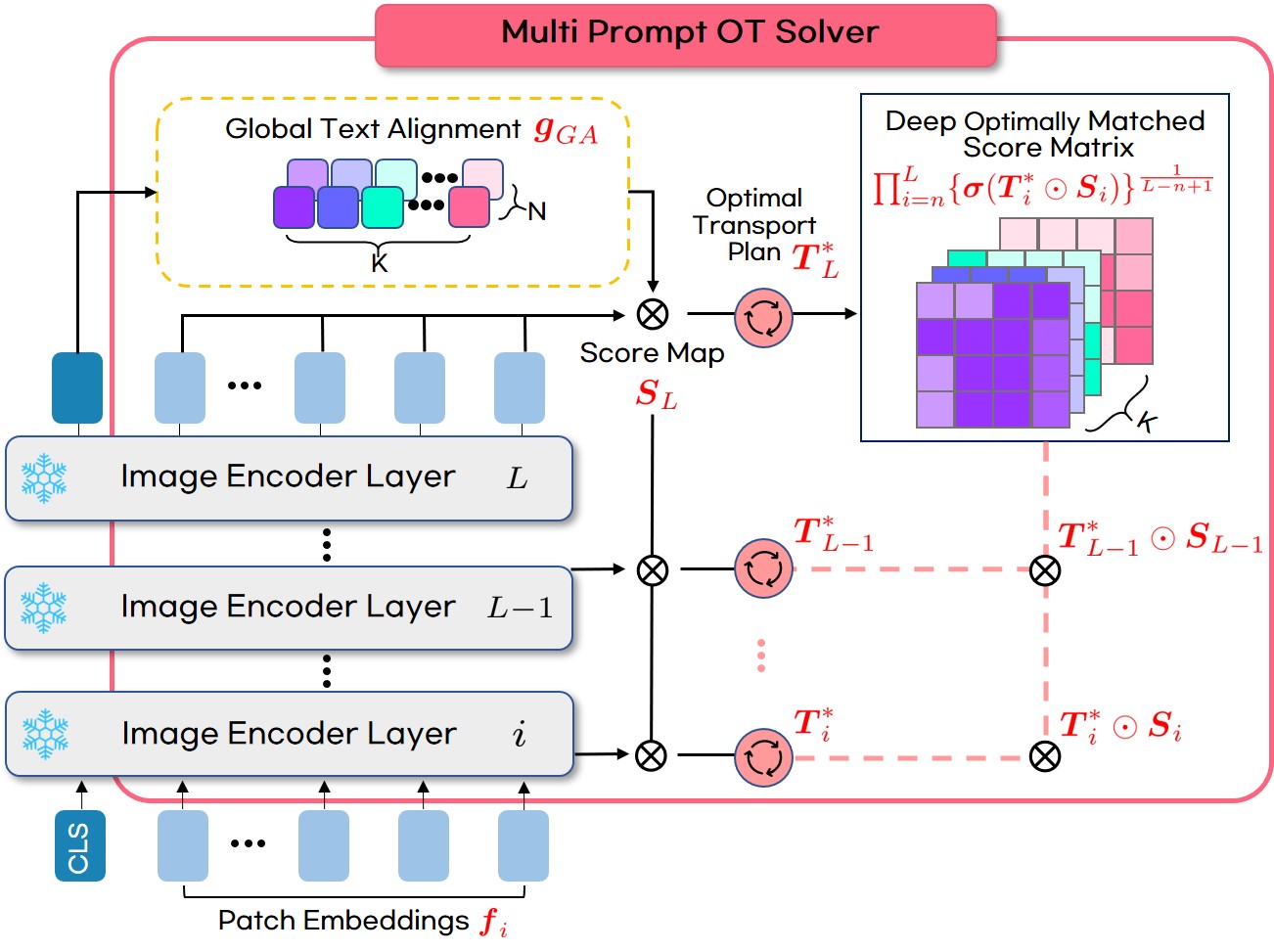}
  \end{center}
  \caption{MPOT solver module. The pixel-text score matrix $\bs{S}_i$ given each image encoder layer is refined through the optimal transport plan $\bs{T}^{\ast}_i$ to yield the the final multi-layer combined score matrix $\bs{S}_i$.}
\end{wrapfigure}
 
Instead of optimizing the transport plan just for the single layer,  our proposed ZegOT leverages intermediate image embeddings from the deep image encoder layers  to yield multi-layer transport plans $\{\bs{T}^{\ast}_i\}_{i=n}^L$,
where $\bs{T}_i \in \mathbb{R}^{H_LW_L \times K N}$ and $n$ denotes the starting layer that comprises the multi-layer transport plans. 
When we combine these optimal transport plans and score maps, 
we apply geometric mean among the multiple transport plans to reflect the knowledge of intermediate score matrices as follows: 
\begin{eqnarray}
\bs{S}^{\ast} = \; \prod^{L}_{i=n} \bs{\sigma}\{\bs{S}^{\ast}_i\}^{\frac{1}{d}}, \, 
\label{eq:deep}
\end{eqnarray}
where $d = L - n + 1$ denotes the depth of incorporating image encoder layers, $\bs{\sigma}$ is the sigmoid function if $n\leq i<L$, and the identity function otherwise~(since intensity range of all the score matrix is (-1,1), we conduct Sigmoid for all the intermediate image embeddings except for the last layer).
Accordingly, in ZegOT, the model prediction $\asty$ can be obtained by pluging \eqref{eq:deep} into \eqref{eq:predict3} and utilizing \eqref{eq:predictfinal}.
Note that Eq.~\eqref{Sinkhorn2} only contains matrix multiplication and exponential operation, \textit{i.e}, the calculations are fully differentiable, thus the gradients can be back-propagated throughout the entire neural network. 

\subsection{Training Procedure and Loss function}
In the training procedure,  we follow the previous SOTA methods\cite{zhou2022zegclip, zhou2022maskclip}, where the entire training schedule is divided into {two phases:
Seen classes-guided learning, and Self-training.}
The details of the training phases are deferred in Supplementary material.

\paragraph{Loss Function} \label{sec:loss}

In this work, we combine three different losses which is similar to previous methods as follows:
\begin{align}
&\mathcal{L}_{\text{seg}} = \lambda_{\text{ce}} \mathcal{L}_{\text{ce}} +\lambda_{\text{fc}} \mathcal{L}_{\text{fc}} + \lambda_{\text{dc}} \mathcal{L}_{\text{dc}}, 	\quad \mathcal{L}_{\text{total}} = 
\mathcal{L}_{\text{seg}}(\mby,\gty) + \mathcal{L}_{\text{seg}}(\mbty, \gty)
\label{losses}
\end{align}
where $\mathcal{L}_{\text{seg}}$ denotes the segmentation loss combining different three losses, $\mathcal{L}_{\text{ce}}$, $\mathcal{L}_{\text{fc}}$ and $\mathcal{L}_{\text{dc}}$ are the cross entropy loss, the focal loss, and the dice loss, with $\lambda_{\text{ce}}$, $\lambda_{\text{fc}}$, and  $\lambda_{\text{dc}}$ as corresponding hyper-parameters, respectively.  
 $\mby \in \mathbb{R}^{HW \times K}$ and $\mbty \in \mathbb{R}^{HW \times K}$ are the predictions of our model which is defined by~\eqref{eq:predict3}, and $\gty \in \mathbb{R}^{HW \times K}$ is the ground-truth label. The details of the loss function are described in Supplementary material. 



\begin{table*}[h!]
\caption{Quantitative comparison of zero-shot semantic segmentation performance of ZegOT with baseline methods on PASCAL VOC 2012,   PASCAL Context, and COCO-Stuff 164K datasets.}
\label{tab_main}
\begin{center}
\resizebox{0.97\linewidth}{!}{
\begin{tabular}{lccccccccc}

\toprule
\multirow{2}{*}{Methods}  & \multicolumn{3}{c}{PASCAL VOC 2012} & \multicolumn{3}{c}{PASCAL Context} & \multicolumn{3}{c}{COCO-Stuff164K} \\
\cmidrule(lr){2-4} \cmidrule(lr){5-7} \cmidrule(lr){8-10} 
 & mIoU (U) & mIoU (S) & hIoU  & mIoU (U) & mIoU (S) & hIoU  & mIoU (U) & mIoU (S)  & hIoU  \\
\cmidrule(r){1-1}\cmidrule(r){2-3}	\cmidrule(r){4-4} \cmidrule(r){5-6} \cmidrule(r){7-7} \cmidrule(r){8-9} \cmidrule(r){10-10} 


    CaGNet~\cite{gu2020cagnet} &  30.3 & 78.6 &43.7 & 16.3 & 36.4  & 33.5 & 13.4 & 35.6  &  19.5  \\
    SPNet~\cite{xian2019semantic}    & 25.8  & 77.8 & 38.8 & - & - & - &  26.9 &  34.6   &  30.3  \\
    STRICT~\cite{pastore2021strict}    & 35.6 & 82.7 & 49.8 & - & - & - &  30.3  &  35.3   &  32.6  \\
    zsseg~\cite{xu2021zsseg} & 78.1  & 79.2 & 79.3 & - & - & - &  43.6  &  38.1  &  41.5  \\
    MaskCLIP+~\cite{zhou2022maskclip} & 88.1 & 86.1 & 87.4 & 66.7 & 48.1 & 53.3 & 54.7& 39.6 & 45.0  \\
    ZegCLIP~\cite{zhou2022zegclip}  & 87.3 & \textbf{92.3} & 91.1 & 68.5 & 46.5 & 55.6 & \textbf{59.9} & \textbf{40.6}  & \textbf{48.4}  \\
    FreeSeg\cite{qin2023freeseg} & 82.6 & {91.8} & 86.9 & - & - & - & 49.1 & {42.2}  & 45.3  \\
    MVP-SEG+\cite{guo2023mvp} & 87.4 & 89.0 & 88.0 & 67.5 & 48.7 & 54.0 & 55.8 &  39.9  & 45.5  \\
\cmidrule(r){1-1}\cmidrule(r){2-3}	\cmidrule(r){4-4} \cmidrule(r){5-6} \cmidrule(r){7-7} \cmidrule(r){8-9} \cmidrule(r){10-10} 
    {{ZegOT (Ours)}} & \textbf{90.9} & {91.9}  & \textbf{91.4} & \textbf{72.5}  & \textbf{50.5} & \textbf{59.5} & {59.2} & {38.2} & {46.5}   \\
    \bottomrule	
        
\end{tabular}
		}
\end{center}
	\vskip -0.1in
\end{table*}

\section{Experiments}

\subsection{Dataset and Evaluation Metric}
\paragraph{Dataset}
To evaluate the effectiveness of our proposed method, we carry out extensive experiments on three challenging datasets: PASCAL VOC 2012~\cite{everingham2012pascal}, PASCAL  Context~\cite{mottaghi2014role}, and COCO-Stuff164K~\cite{caesar2018coco}.  To fairly compare with previous methods~\cite{bucher2019zero,xu2021zsseg,ding2022zegformer,zhou2022maskclip,zhou2022zegclip}, we follow the identical protocol of dividing seen and unseen classes for each dataset. 
To evaluate the effectiveness of our proposed method, we carry out extensive experiments on three challenging datasets: PASCAL VOC 2012~\cite{everingham2012pascal}, PASCAL  Context~\cite{mottaghi2014role}, and COCO-Stuff164K~\cite{caesar2018coco}.  To fairly compare with previous methods~\cite{bucher2019zero,xu2021zsseg,ding2022zegformer,zhou2022maskclip,zhou2022zegclip}, we follow the identical protocol of dividing seen and unseen classes for each dataset. 
The dataset details are described in Supplementary material.

\vspace{-0.2em}
\paragraph{Evaluation Metric}
By following previous works, we measure the mean of class-wise intersection over union (mIoU) on both seen and unseen classes, indicated as mIoU(S) and mIoU(U), respectively. We adopt the harmonic mean IoU (hIoU) of seen classes and unseen classes as a major metric. More details of the definition are deferred to Appendix. We also report the pixel-wise classification accuracy (pAcc) in Supplementary material. 

\subsection{Implementation Details}
We implement the proposed method on the open-source toolbox MMSegmentation ~\cite{mmseg2020} \footnote{\url{https://github.com/open-mmlab/mmsegmentation}} 
and conduct our algorithm on {at most} 8 NVIDIA A100 GPUs with batch size of 16.
We adopt the pre-trained CLIP ViT-B/16 model\footnote{\url{https://github.com/openai/CLIP}} as the frozen encoder module and 
adopt FPN which is equipped with atrous spatial pyramid pooling (ASPP) \cite{chen2017deeplab} module as the image decoder 
for all the experiments. Furthermore, we set $\lambda$ to 0.2 in~\eqref{eq:predictfinal}.
Further details are deferred to Supplementary material. We will release our source code for reproduction.


\subsection{Experimental Results}

\paragraph{Zero-shot Semantic Segmentation}
Figure~\ref{fig_seg} shows the qualitative zero-shot segmentation performance of our ZegOT and the baseline method. Our ZegOT segments semantic objects most accurately, and shows superior performance on sectioning semantic boundaries of unseen objects compared to the previous SOTA methods (see the red arrows). {More visual results are provided in Supplementary material.}
Quantitative results are also presented in Table~\ref{tab_main}, and our ZegOT achieves the SOTA performance for most datasets. 
{To further evaluate the generalization capabilities of ZegOT, we performed a cross-domain experiment between COCO-Stuff 164K and PASCAL Context datasets, as shown in Table~\ref{table_gen}. It demonstrate the superior performance of ZegOT compared to the previous SOTA method in both experimental setting and capability of ZegOT for domain generalization. }

\begin{figure*}[!t]
\vskip 0.1in
\begin{center}
\includegraphics[width=0.94\linewidth]{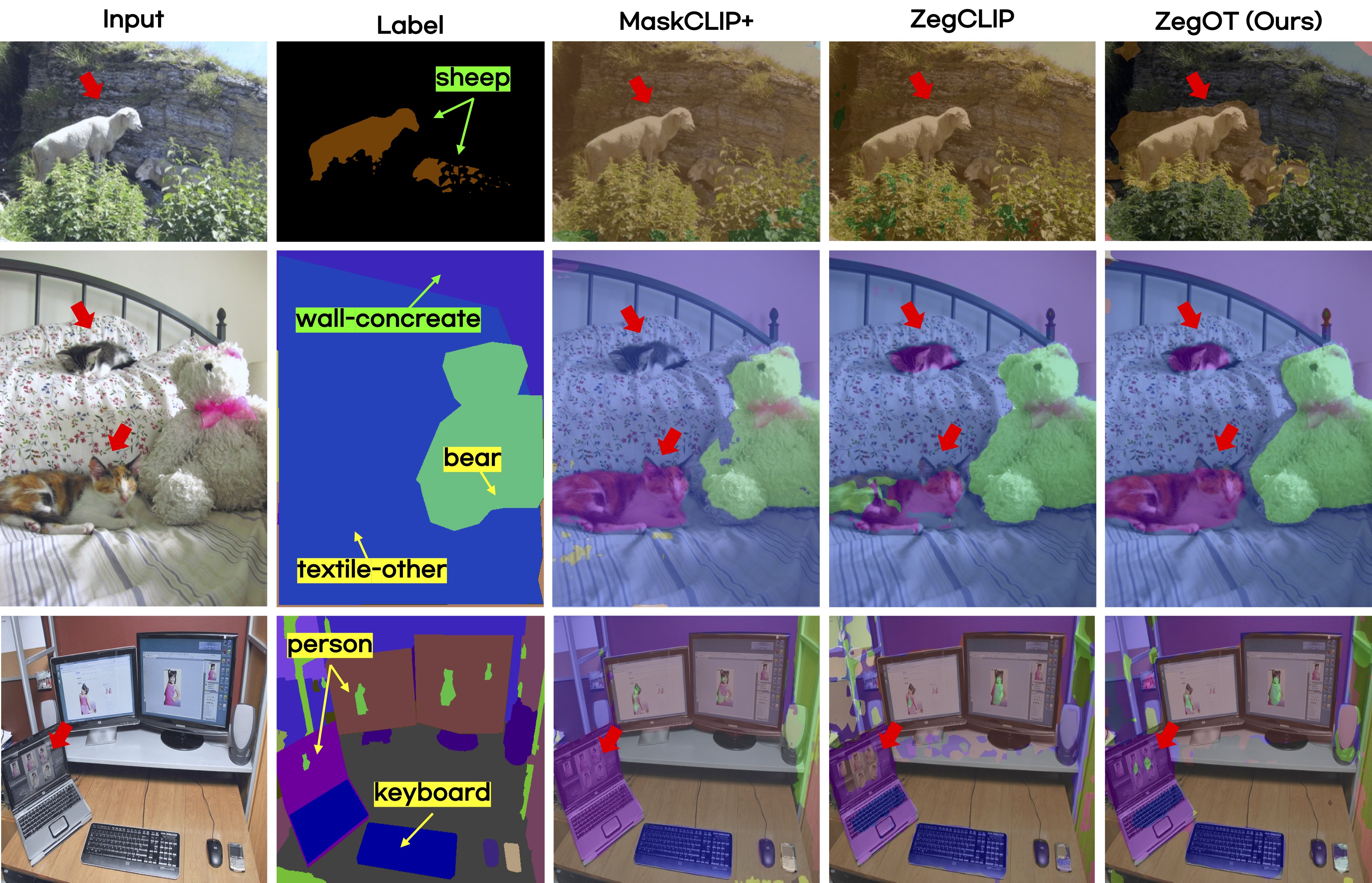}
\caption{Qualitative segmentation performance comparison with the previous SOTA models. The \colorbox{green}{green} tag indicates unseen classes, while the \colorbox{yellow}{yellow} tag indicates seen classes.} 
\label{fig_seg}
\end{center}
\vskip -0.1in
\end{figure*}



\begin{figure}[ht] 
\begin{minipage}{0.64\linewidth}
    \centering
    \includegraphics[width =\linewidth]{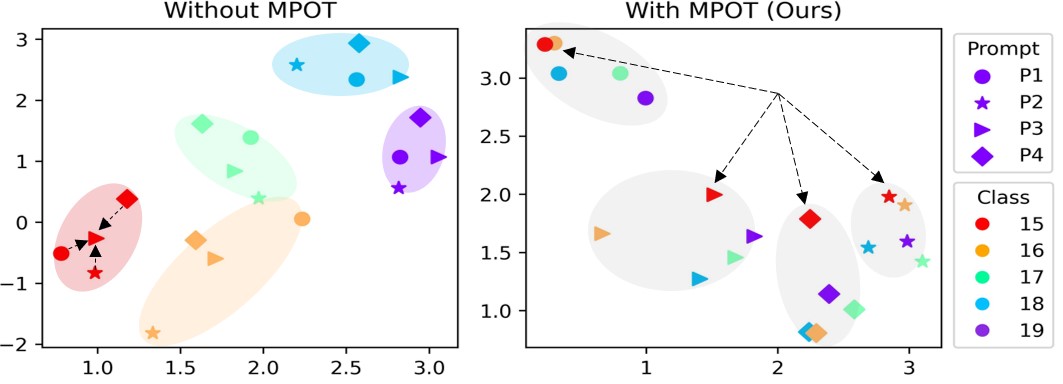}
    \vspace{-0.1cm}
    \captionof{figure}{Analysis of the learned text prompts distribution.}
    \label{fig_tsne}
    \vskip -0.1in
\end{minipage}
\hspace{0.1cm}
\begin{minipage}{0.34\linewidth}
    \centering
    \resizebox{1\linewidth}{!}{
    \begin{tabular}{ccc}
    \toprule
    \multirow{2}{*}{Method} & \multirow{2}{*}{Source} &\multicolumn{1}{c}{Target} \\
    \cmidrule(r){3-3}  
     &  & Context\\ 
    \cmidrule(r){1-1} \cmidrule(r){2-2} \cmidrule(r){3-3}  
     ZegFormer~\cite{ding2022zegformer} &\multirow{3}{*}{COCO} & 29.3 \\ 
     ZegCLIP~\cite{zhou2022zegclip}&  & 41.2  \\
     Ours &  &  \textbf{41.3}  \\
    \midrule 
     Method  &  Source & COCO \\ 
    \cmidrule(r){1-1} \cmidrule(r){2-2} \cmidrule(r){3-3}         
    ZegCLIP~\cite{zhou2022zegclip} & \multirow{2}{*}{Context} & 16.7 \\ 
     Ours & &  \textbf{17.8} \\
    \bottomrule
    \end{tabular}
    }
    \captionof{table}{Generalization performance of the open vocabulary segmentation on cross dataset.} 
    \label{table_gen}
\end{minipage}
\end{figure}


\paragraph{MPOT-driven Text Prompt-Image Alignment}
To explore the origin of the superior segmentation performance of our ZegOT, we conduct an extensive analysis to examine the distribution of learned text prompts. {Firstly, we visualize each text prompt-related score matrix in Figure~\ref{fig_main}. We further} visualize the learned distribution of text embeddings by utilizing t-SNE projection \cite{hinton2002stochastic}, as illustrated in Figure~\ref{fig_tsne}. For comparison, we set a baseline which basically shares ZegOT structure but the MPOT module is ablated. In the baseline method without MPOT, the learned text prompts $\bs{P}$ show tightly grouped distribution within their respective class clusters (indicated as different colors in the left image, respectively). The result suggests that the learned text prompts lack the ability to attend on different semantic context but converge to represent similar semantic context. In contrast, our ZegOT with MPOT produces dispersed distributions, with even cohesion within corresponding prompt clusters (indicated as gray circles in the right image in Figure~\ref{fig_tsne}, respectively).

\subsection{Ablation Studies and Empirical Analyses}

\begin{table}[t!]
\caption{Ablation studies for the main proposed component and validation for matching approach.}
\label{tab:ablation}
\begin{center}
\resizebox{1\linewidth}{!}{
\begin{tabular}{cccccccccccc}

\toprule
\multirow{2}{*}{Method} & \multirow{2}{*}{MPOT} &  \multirow{2}{*}{Matching method}  & \multicolumn{3}{c}{PASCAL VOC 2012} & \multicolumn{3}{c}{PASCAL Context} \\ 
 \cmidrule(lr){4-6} \cmidrule(lr){7-9}

& ~ & ~  & mIoU (U) &  mIoU (S) & hIoU  & mIoU (U) &  mIoU (S) & hIoU  \\
 \cmidrule(r){1-1} \cmidrule(r){2-3} \cmidrule(r){4-5} \cmidrule(r){6-6}  \cmidrule(r){7-8} \cmidrule(r){9-9}
 (a) & \xmark & - & 85.4 & 91.9 & 88.5 & 65.0 & \textbf{51.0}  & 57.1  \\
 \cmidrule(r){1-1} \cmidrule(r){2-3} \cmidrule(r){4-5} \cmidrule(r){6-6}  \cmidrule(r){7-8} \cmidrule(r){9-9}
Ours & \checkmark &  Sinkhorn \cite{cuturi2013sinkhorn} & \textbf{90.9} & \textbf{91.9} & \textbf{91.4}   & \textbf{72.5}  & {50.5} & \textbf{59.5} \\
 \cmidrule(r){1-1} \cmidrule(r){2-3} \cmidrule(r){4-5} \cmidrule(r){6-6}  \cmidrule(r){7-8} \cmidrule(r){9-9}
(b) & \xmark & Bipartite \cite{kuhn1955hungarian} & 82.3 & 89.0 & 85.7 & {57.6} & {46.7} & {51.6}  \\
(c)& \xmark & Self-attention \cite{vaswani2017attention} & 84.4 & 91.4 & 87.7 & 59.3 & 50.6 & 54.6   \\
\bottomrule
\end{tabular}
}
\end{center}
\vskip -0.1in
\end{table}


In our proposed framework, the most crucial point is the optimal alignment of the vision-language score map, specifically mapping each pixel to the multiple text prompts. {To demonstrate the effectiveness of our proposed MPOT solver, we conduct ablation studies as reported in Table~\ref{tab:ablation}. (a) Our model without MPOT shows inferior performance on unseen classes with margins of 5.5$\%$ and 7.5 $\%$ mIoU for each PASCAL VOC 2012 and Context dataset, respectively. 
We further alter our MPOT with two comparative matching methods: (b) Hungarian algorithm~\cite{kuhn1955hungarian} and (c) Self-attention algorithm~\cite{vaswani2017attention}. The Hungarian algorithm, which refers to Bipartite Matching, performs an one-to-one assignment where each pixel is assigned to only one of text prompts. In contrast, we chose the Sinkhorn algorithm~\cite{cuturi2013sinkhorn} for our ZegOT, enabling one-to-many assignment where each pixel can be partially assigned to all the multiple text prompts. 
In the results presented in Table~\ref{tab:ablation}, it is evident that the Sinkhorn algorithm outperforms the bipartite method by a margin of 5.7$\%$ hIoU for the PASCAL VOC 2012, and 7.9 $\%$ hIoU for the PASCAL context. 
The one-to-one assignment, where only one prompt can be activated per pixel, overlooks significant overlapping semantic features for each text prompt. In contrast, our one-to-many assignment preserves meaningful semantic features for each text prompt. 
{To ensure a fair comparison, when considering self-attention matching, we replace the MPOT Solver in our ZegOT framework with lightweight transformer blocks that consist of a single-head self-attention module}.  
When compared to the Self-attention method, our ZegOT achieves higher unseen classes performance with margins of 6.5$\%$ and 13.2 $\%$ mIoU for the respective datasets.
The Self-attention method inherently} uses learnable projection layers, which allows the model to adapt to task-specific knowledge but prevents it from preserving the frozen VLM knowledge. 
{The results} demonstrate that our ZegOT achieves the best unseen classes performance by optimally transporting multiple text prompts for dense prediction in the zero-shot segmentation setting.

\section{Conclusion}
In this work, we proposed ZegOT, a novel framework for zero-shot semantic segmentation, which thoroughly leverages the aligned vision-language knowledge of a frozen visual-language model by optimizing the multiple text prompts 
We also incorporated optimal transport theory into our framework to train multiple text prompts for representing different semantic properties of visual inputs. 
We demonstrated that our ZegOT outperformed the SOTA zero-shot segmentation approaches on various benchmark datasets. 
Our in-depth analyses also confirmed that our ZegOT effectively delivered performance gains on both seen and unseen classes. 

\textbf{Limitations}
Our model performance is highly dependent on the CLIP model knowledge. Specially, the class name-driven segmentation method can be failed, unless the CLIP model is pre-trained with the target class-related image-text pairs. 

\textbf{Negative Societal Impacts}
Our paper has a potential privacy issue in providing visual results of the person class. To avoid the issue, every person in all the figures are properly anonymized by covering the faces.


\clearpage
\newpage
\newpage

\appendix

\section{Zero-shot Segmentation Training Process.} \label{sec_phase}

By following the general transductive zero-shot semantic segmentation (ZS3) setting \cite{zhou2022maskclip, zhou2022zegclip}, we divide the entire training procedure into two phases, as described in Algorithm \ref{algoritm_phases1}.

\begin{algorithm}
\caption{ZegOT Pseudo-code}\label{algoritm_phases1}
\label{alg:phase1}
{\bfseries Input:} ZegOT model $Z_t$ at iteration $t$, The subset of seen classes $\mc{C}_S$ and unseen classes $\mc{C}_U$, $\textit{e.g,}$ $\mc{C}_S \cap \mc{C}_U = \emptyset$. The training dataset $D = \{(x,\gty)| x \in \mathcal{X}, \gty_{hw}\in \mc{C}_S\}$, the training iterations for seen class-guided learning $T_g$, and self-training $T_s$;
\\
{\bfseries Phase 1: Seen class-guided learning}\\
\For{$t=1, 2, \cdots, T_g$ }{
	$\mbf{Y}, \mbty \leftarrow$ model prediction from $Z_{t}(x)$;\\
    $\mathcal{L}_{} \leftarrow \mathcal{L}_{\text{seg}}(\mbf{Y}, \gty) + \mathcal{L}_{\text{seg}}(\mbty,\gty)$; 
    \\
    $Z_{t+1}\leftarrow$ AdamW model parameter update;}

	
{\bfseries Phase 2: Self-training}

\For{$t=T_{g} +1,T_{g} + 2, \cdots, T_{g}+T_s$}
{
$\mbty \leftarrow$ model prediction from $Z_{t-1}(x)$; \\
\If{$\gty_{hw} \not\in \mc{C}_S$}
{$\gty_{hw} = \underset{\bs{c} \in \mc{C}_U}{\arg{\max}} \ (\mbty_{hw} = \bs{c}|x)$;}
$\mbf{Y}, \mbty \leftarrow$ model prediction from $Z_{t}(x)$; \\
$ \mathcal{L} \leftarrow \mathcal{L}_{\text{seg}}(\mbf{Y}, \gty) + \mathcal{L}_{\text{seg}}(\mbty, \gty)$; \\ 
$Z_{t+1}\leftarrow$ AdamW model parameter update;
}   
\end{algorithm}

\paragraph{Phase 1. Seen Class-guided Learning}
The model is trained utilizing the ground truth label $\gty$, which only contains pixel-wise labels $\gty_{hw}$ for seen classes $\mc{C}_S$, and the rest labels are ignored for calculating losses. 
{$T_g$ is fixed as 20\% of total training iterations.}


\paragraph{Phase 2. Self-training}
For the rest of the training iterations, the model self-generates each pixel values of the ground truth label $\gty_{hw}$ that does not belong to $\mc{C}_S$, at every training iteration. 
{For large-scale datasets with complex classes, e.g., PASCAL Context and COCO-Stuff164K, we firstly fix $t$ of $Z_{t-1}$ as $T_{g}$ to ensure stability in the label generation process, and utilize $Z_{{T_{g}}-1}$ until $t$ reaches 50\% of total training iterations. For the rest of the self-training period, $t$ of $Z_{t-1}$ is updated simultaneously. }

\section{Additional Explanation of Component Architecture}
\label{sec:additonal}
 \paragraph{Global Text Alignment (GTA)}
 The local image embeddings $\bs{f}_L$ inherently contain the text-image aligned knowledge of CLIP. 
However, when pre-training, the global image embedding $\bs{\bar{f}}_L$  and the text embeddings  $\bs{g}$ comprise the cosine similarity score which is maximized through contrastive learning so that $\bs{\bar{f}}_L$ contains richer pixel-text aligned information than $\bs{f}_L$. 
Thus, we further exploit $\bs{\bar{f}}_L$ in the GTA module to compute the score matrix $\bss \in \mathbb{R}^{H_LW_L \times KN}$ by employing an idea of relation descriptor method \cite{zhou2022zegclip} in \eqref{eq:score}.

\paragraph{Deep Local Feature Alignment (DLFA)}
 Feature Pyramid Network (FPN) \cite{lin2017fpn} is a common choice for segmentation decoder, 
where high-level semantic features from deep encoder layers comprise a latent feature for the following image decoder.
 Our ZegOT also adopts FPN,  but we fully align the extracted intermediate local embeddings from the frozen image encoder with the globally aligned text embedding $\bs{g}_{GA}$. In other words, we extend the idea of GTA to all the intermediate local embeddings, $\textit{i.e.,}$ $\{\bss_i\}_{i=1}^{L}$, where $\bss_i \in \mathbb{R}^{H_LW_L \times KN}$ is the $i$-th score matrix.
Through the simple arithmetic calculation, the entire hidden image embeddings extracted from the frozen CLIP encoder can be deeply aligned with both the global image and text embeddings without any further learnable parameters.

\vspace{-0.2em}

\section{Analysis on Deep Local Feature Alignment (DLFA)}

To demonstrate the effectiveness of the proposed DLFA, we conduct in-depth analysis comparing the strength of feature alignment with or without DLFA on PASCAL VOC 2012 dataset.
Figure~\ref{DLFA} shows the bar plot of average text-pixel alignment given a specific class name over the image encoder layer index.
To calculate the strength of feature alignment, we extract the score matrices $\{\bss_i|_{i=1}^{L}\}$ for a certain class name (\textit{e.g.}, dog) from the trained model and calculate the average $\bss_i$ along entire pixels for each image encoder layer.
We find that the strength of feature alignment with DLFA is much higher compared to that computed through the model without DLFA in almost layers. 
In particular, the average value of text-pixel alignment significantly increases at the earlier layers and the final layer. The result confirms that our model with DLFA effectively exploits the activated alignment between the two language-text modalities.

\begin{figure}[t]
\vskip 0.1in
\begin{center}
\includegraphics[width=0.6\linewidth]{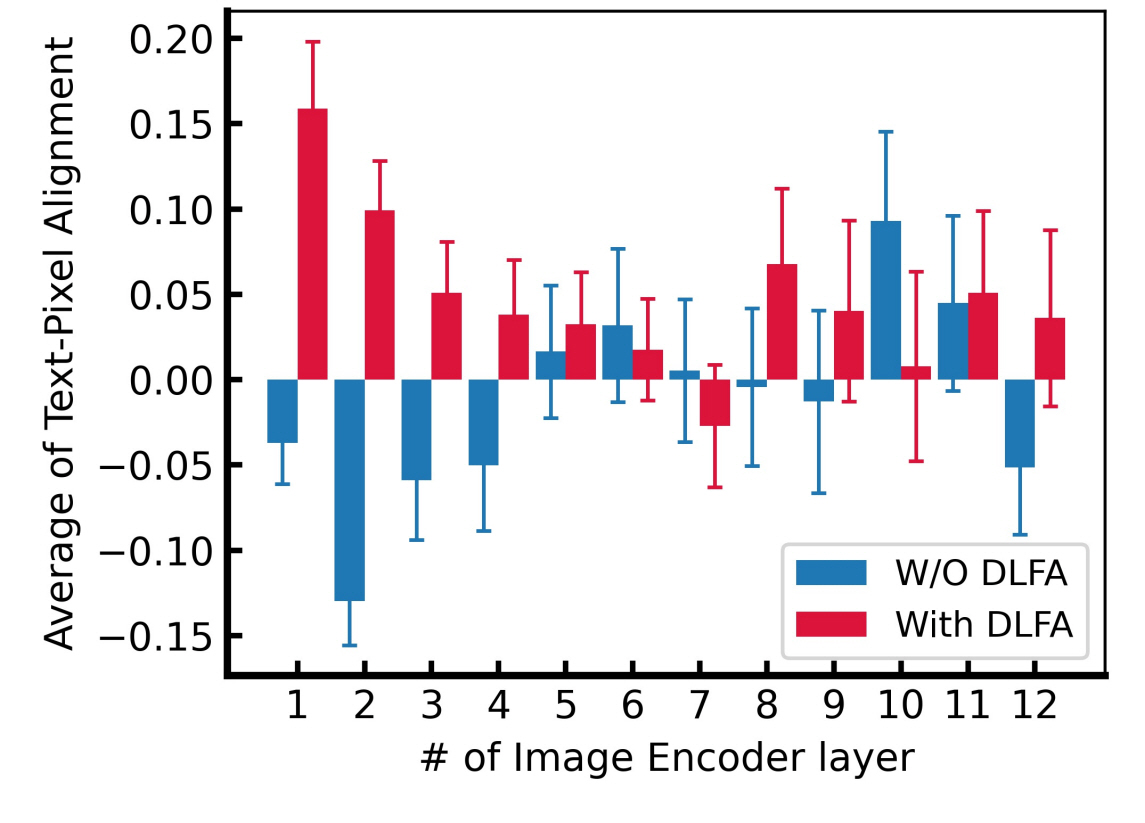}
 \vspace{-0.35cm}
\caption{Bar plots of the average pixel-text alignment versus the image encoder layer index given a specific class name with or without Deep Local Feature Alignment (DLFA).}
\label{fig_dlfa}
\label{DLFA}
\end{center}
\vskip -0.1in
\end{figure}

\section{Further Ablation Studies on Visual-Text Prompts Alignment.}
\label{appendix:pAcc}
We conduct ablation experiments to evaluate the alignment of visual-text prompts as shown in Table~\ref{tab_appn_ablation}. (a) To investigate the proposed framework's synergistic effect, we replace the learnable text prompt with hand-crafted prompts. Our model, equipped with all proposed components, achieves superior performance with margins of 10.6 $\%$ hIoU and 7.9 $\%$ hIoU for PASCAL VOC 2012 and PASCAL Context dataset, respectively. (b) We further ablate the MPOT module and the results show no significant difference to those of (a), which implies that our MPOT module can boost the segmentation performance when incorporated with learnable text prompts as shown in Table~\ref{tab_appn_ablation}-Ours.
(c) Lastly, as discussed in the Section~\ref{sec:additonal}, we examine the impact of the global text-alignment (GTA) layer. Interestingly, when using hand-crafted prompts, removing the GTA layer leads to a significant decrease in performance with margins of 50.3 $\%$ hIoU, and 31.4 $\%$ hIoU for each PASCAL VOC 2012, and PASCAL Context dataset, respectively.  These findings validate the effectiveness of the proposed framework, which incorporates learnable text prompts with the GTA layer and the MPOT solver, by optimally aligning text-visual embeddings to achieve the best performance in the zero-shot segmentation task.

\begin{table}[!h]
\caption{Ablation studies of the network components.}
\label{tab_appn_ablation}
\begin{center}
\resizebox{1\linewidth}{!}{
\begin{tabular}{ccccccccccccc}

\toprule
\multirow{2}{*}{Method} & \multirow{2}{*}{MPOT} & \multirow{2}{*}{GTA}  &  \multirow{2}{*}{Text prompt} & \multicolumn{3}{c}{PASCAL VOC 2012} & \multicolumn{3}{c}{PASCAL Context} \\ 
 \cmidrule(lr){5-7} \cmidrule(lr){8-10}

& ~ & ~ & ~ & mIoU (U) &  mIoU (S) & hIoU  & mIoU (U) &  mIoU (S) & hIoU  \\
 \cmidrule(r){1-1} \cmidrule(r){2-4} \cmidrule(r){5-6} \cmidrule(r){7-7}  \cmidrule(r){8-9} \cmidrule(r){10-10}

Ours & \checkmark & \checkmark &  Learnable   & \textbf{90.9} & \textbf{91.9} & \textbf{91.4}  
& \textbf{72.5}  & \textbf{50.5} & \textbf{59.5} \\
 \cmidrule(r){1-1} \cmidrule(r){2-4} \cmidrule(r){5-6} \cmidrule(r){7-7}  \cmidrule(r){8-9} \cmidrule(r){10-10}

(a) & \checkmark & \checkmark &  Hand-crafted & 78.1 & 83.7 & 80.8  & 54.9 & 33.5 & 41.6 \\
(b) & \xmark & \checkmark  & Hand-crafted & 77.8 & 83.7 & 85.4 & 54.7 & 33.4 & 41.4 \\
(c) & \xmark & \xmark  & Hand-crafted & 33.9 & 36.3 & 35.1 & 15.2 & 7.5 & 10.0  \\

\bottomrule
\end{tabular}
}
\end{center}
\vskip -0.1in
\end{table}

\section{Component Analysis.}
\label{appendix:compo}

To study the effects of hyper-parameters of network components on the zero-shot segmentation performance, we conduct the component analysis in Table~\ref{tab:Component} with varying hyper-parameters including: MPOT layers ($L$), number of text prompts ($N$), and context length ($C$). Firstly, we investigate the effect of incorporating layers $L$ for the MPOT solver, $\textit{i.e.,}$ layers of the frozen image encoder that inserted to MPOT. 
Although inserting multiple layers to MPOT boosts performance, it also causes performance trade-off between seen and unseen classes segmentation. We find that introducing the MPOT with 8 to 12-th layers of the image encoder achieves the best performance, which becomes the default setting of ZegOT for the entire experiments.
We further explore the effect of the depth of MPOT solver, $\textit{i.e.,}$ ranges of the frozen image encoder layers that inserted to MPOT.
Further, we observe the segmentation performance by varying the total number $N$ of the multiple text prompts. We empirically find that ZegOT performs the best when $N \geq 4$, but the hIoU performance drops when $N$ is increased over certain threshold, which implies that fewer text prompts are insufficient to learn comprehensive semantic features, whereas too many text prompts complicate the optimal transport matching process. {Lastly, we empirically find that the large context length ($C$) is limited to take significant effect on segmentation performance, \textit{e.g,} when $C = 16$ the performance on seen classes is the best, but the performance on unseen classes drops. Since we consider the hIoU as a major metric, we adopt $C = 8$ as the default setting.}


\begin{table}[h!]

\caption{Component analysis of hyper-parameters. $\#$ denotes the hyper-parameters of configurations. Checkmark $\checkmark$ indicates the default configuration of ZegOT.}
\label{tab:Component}
\vskip 0.1in
\begin{center}
\resizebox{0.6\linewidth}{!}{
\begin{tabular}{clccccc} 
	
\toprule
\multirow{2}{*}{Components} & \multirow{2}{*}{\#}  & \multicolumn{4}{c}{PASCAL VOC 2012} \\ 
\cmidrule(lr){3-6} 

& & mIoU (U) & mIoU (S) & hIoU & Ours \\

\cmidrule(r){1-1} \cmidrule(r){2-2}	\cmidrule(r){3-4} \cmidrule(r){5-5} \cmidrule(r){6-6} 

\multirow{5}{*}{MPOT layers (L)} & 12 & 91.8& 87.3 & 89.5 &  \\
& 10-12 & 91.4& 87.3& 89.3\\
& 8-12 & 90.9& \bf{91.9}& \bf{91.4}& \checkmark \\
& 4-12 & 91.6& 90.2& 90.9 \\
& 1-12 & \bf{92.1} & 86.5 & 89.3 &\\

\cmidrule(r){1-1} \cmidrule(r){2-2}	\cmidrule(r){3-4} \cmidrule(r){5-5} \cmidrule(r){6-6} 

\multirow{3}{*}{Text prompts (N)} & 2 &  82.9 & 91.6 & 87.0 \\
& 4 & \textbf{90.9} & \textbf{91.9} & \textbf{91.4} & \checkmark\\
& 6 & 89.6 & 91.8 & 90.7 \\

\cmidrule(r){1-1} \cmidrule(r){2-2}	\cmidrule(r){3-4} \cmidrule(r){5-5} \cmidrule(r){6-6} 

\multirow{3}{*}{Context lengths (C)} & 8 & \textbf{90.9} & {91.9} & \textbf{91.4} & \checkmark\\
& 16 & 89.6 & \bf{92.2} & 90.9\\
& 32 & 90.3 & 91.9 & 91.1\\

\bottomrule
\end{tabular}
}
\end{center}
\vskip -0.1in
\end{table}

\section{Details of Loss function}
\label{appendix:loss}
As discussed in Section~\ref{sec:loss}, we combine three different losses, including the focal loss based on Cross Entropy (CE) loss, and the dice loss, which are given  by:
\begin{align}
\begin{split}
\mathcal{L}_{\text{ce}} = -\frac{1}{HW} \sum^{HW}_{i=1} \gty_i \log(\phi(\mbf{Y}_i)) \\
    + (1- \gty_i)\log(1-\phi(\mbf{Y}_i)), \\   
\end{split}\\
\begin{split}
\mathcal{L}_{\text{focal}} = -\frac{1}{HW} \sum^{HW}_{i=1}\gty_i (1- \sigma(\mbf{Y}_i))^{\gamma} \log(\sigma(\mbf{Y}_i) )\\
    + (1- \gty_i)\sigma(\mbf{Y}_i)^{\gamma}\log(1-\sigma(\mbf{Y}_i)), \\   
\end{split}\\    
\begin{split}
\mathcal{L}_{\text{dice}} = 1 -\frac{2\sum^{HW}_{i=1}\gty_i \mbf{Y}_i}{\sum^{HW}_{i=1} {\gty_i}^2 + \sum^{HW}_{i=1} {\mbf{Y}_i}^2}, 
\end{split} \\
\end{align}
where $\mbf{Y}$ is the model decoder outputs, $\gty$ is the ground truth label, 
$\sigma(\cdot)$ is Sigmoid operations, 
$\gamma$ is a hyper-parameter to balance hard and easy samples, which is set to 2.
Throughout the entire experiments,
$\lambda_{\text{ce}}$, $\lambda_{\text{focal}}$ and $\lambda_{\text{dice}}$ are set to 1, 20, and 1, respectively. 

\section{Details of Dataset}
\label{appendix:datsset}

We utilize a total of three datasets, $\textit{i.e.,}$ PASCAL VOC 2012~\cite{everingham2012pascal}, PASCAL  Context~\cite{mottaghi2014role}, and COCO-Stuff164K~\cite{caesar2018coco}. 
We divide seen and unseen classes for each dataset, following the settings of previous methods~\cite{bucher2019zero,xu2021zsseg,ding2022zegformer,zhou2022maskclip,zhou2022zegclip}. PASCAL VOC 2012 consists of 10,582 / 1,449 images with 20 categories, for training / validation. The dataset is divided into 15 seen and 5 unseen classes. PASCAL Context is an extensive dataset of PASCAL VOC 2010 that contains 4,996 / 5,104 images for training / test with 60 categories. The dataset is categorized into 50 seen and 10 unseen classes. COCO-Stuff 164K is a large-scale dataset that consists of 118,287 / 5,000 images for training / validation with 171 classes. The dataset is categorized into 156 seen and 15 unseen classes.

\section{Definition of Harmonic mean IoU}
\label{appendix:hIoU}
Following the previous works ~\cite{xu2021zsseg,zhou2022maskclip,zhou2022zegclip}, we define harmonic mean IoU (hIoU) among the seen and unseen classes as:
\begin{align}
    \text{hIoU} =\frac{2 * \text{mIoU(S) +mIoU(U)}}{\text{mIoU(S)+mIoU(U)}}
\end{align}

\section{Implementation Detail}
\label{appendix:implement}
 We further declare the implementation detail for our work. 
 Input image resolution is set as 480$\times$480 for PASCAL Context, and 512$\times$512 for the rest of the datasets. The context length is set to 8 and the weight $\lambda$ for the score matrix~\eqref{eq:predictfinal} is set to 0.2. We choose the lightest training schedule, which is 20K / 40K / 80K for PASCAL VOC 2012 / PASCAL Context / COCO-Stuff-164K.


\section{Additional Visual Results.}
\label{appendix:fig}
We provide additional visual results. Figure~\ref{fig_seg_context} and Figure~\ref{fig_seg_coco} show qualitative zero-shot segmentation performance of our ZegOT for PASCAL Context and COCO-Stuff164K datasets. 
For PASCAL Context dataset, we reproduce the segmentation results using a publically available model weight \cite{zhou2022maskclip}. Our ZegOT segments semantic objects most accurately of unseen objects compared to the previous SOTA methods in Figure~\ref{fig_seg_context} (see the red arrows).
For COCO-Stuff164K dataset, we reproduce the segmentation results using publically available weights \cite{zhou2022maskclip, zhou2022zegclip}.



\begin{figure*}[!h]
\vskip 0.1in
\begin{center}
\includegraphics[width=0.97\linewidth]{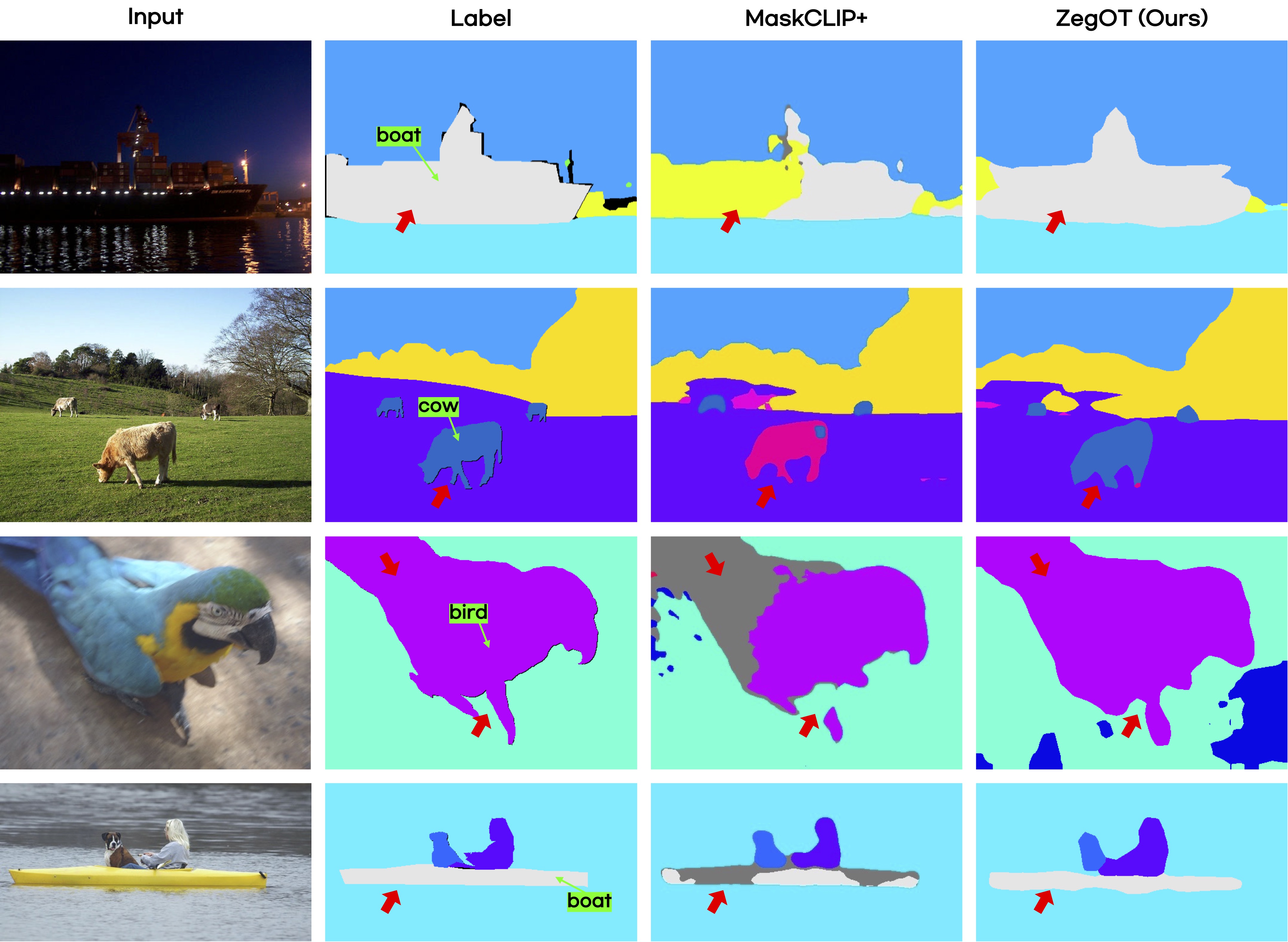}
\caption{Qualitative zero-shot segmentation results of PASCAL Context datasets. The \colorbox{green}{green} tag indicates unseen classes. }
\label{fig_seg_context}
\end{center}
\vskip -0.1in
\end{figure*}


\begin{figure*}[!h]
\vskip 0.1in
\begin{center}
\includegraphics[width=0.97\linewidth]{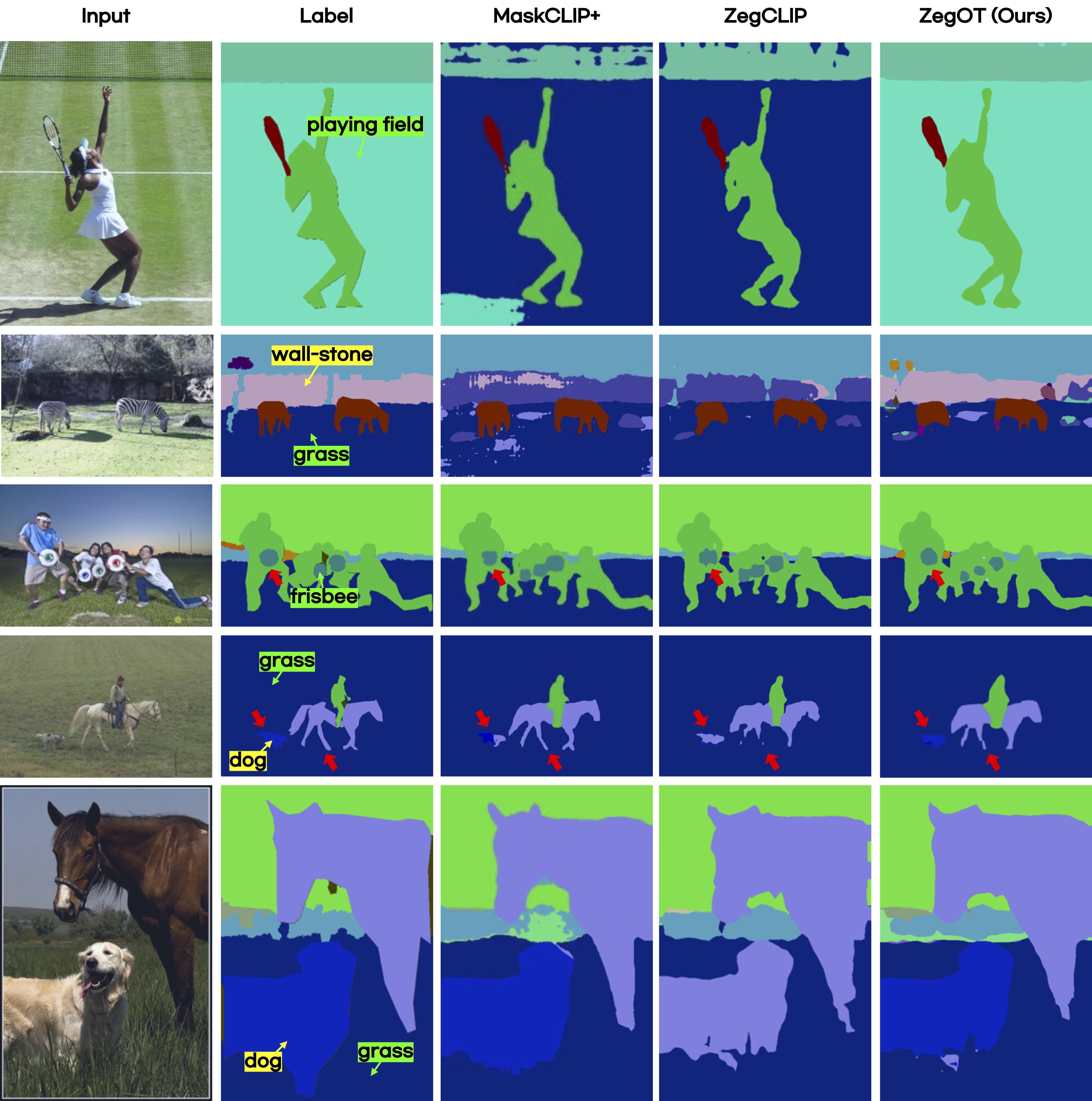}
\caption{Qualitative zero-shot segmentation results of COCO-Stuff164K dataset. The  \colorbox{yellow}{yellow} tag indicates seen classes, while the \colorbox{green}{green} tag indicates unseen classes.}
\label{fig_seg_coco}
\end{center}
\vskip -0.1in
\end{figure*}



\end{document}